# Tinny CNN for feature point description for document analysis: approach and dataset

*A. Sheshkus [1,2,3], A. Chirvonaya [4,3], V.L. Arlazarov [2,3]*
*[1]Moscow Institute for Physics and Technology,*
*141701, Russia, Moscow Region, Dolgoprudny, Institutskiy per., 9,*
*[2] Institute for Systems Analysis, Federal Research Center "Computer Science and*
*Control" of Russian Academy of Sciences, 117312, Moscow, Russia, pr. 60-letiya Oktyabrya, 9,*
*[3] Smart Engines Service LLC, 117312, Moscow, Russia, pr. 60-letiya Oktyabrya, 9,*
*[4] National University of Science and Technology "MISIS", 119049, Moscow, Russia, Leninskiy prospect, 4.*

*Abstract*

In this paper, we study the problem of feature points description in the context of document analysis and template matching. Our study shows that the specific training data is required for the task especially if we are to train a lightweight neural network that will be usable on devices with limited computational resources. In this paper, we construct and provide a dataset with a method of training patches retrieval. We prove the effectiveness of this data by training a lightweight neural network and show how it performs in both documents and general patches matching. The training was done on the provided dataset in comparison with HPatches training dataset and for the testing we use HPatches testing framework and two publicly available datasets with various documents pictured on complex backgrounds: MIDV-500 and MIDV-2019.

<u>*Keywords*</u>: feature points description, metrics learning, training dataset.

<u>*Citation:*</u> **Sheshkus A, Chirvonaya A, Arlazarov VL.** Tinny CNN for feature point description for document analysis: approach and dataset. Computer Optics 20XX; 4X(X): XXX-YYY. DOI: 10.18287/2412-6179-CO- editorial index.

<u>*Acknowledgements*</u>: This work was supported by the Russian Foundation for Basic Research (projects 18-29-26033 and 19-29-09064)

## Introduction

The image description is a very important part of computer vision in modern science. The algorithms that somehow build a representation for the object are required in many scopes from image tagging and annotation for medical [1] or other purposes [2] to face verification [3]. The purpose of these methods is to transform an object (image, image patch, signal) into a vector of values. The essential property of these methods is to yield comparable vectors: the distance between these vectors must be small for the representations of the same/similar object and big for the representations of the different objects. In the scope of the document understanding and recognition these algorithms are also playing an important role. They are used for document template matching [4], forensics checks [5] and even for recognition of the characters [6].

Two main types of descriptors are used: binary and floating point. The main advantage of the binary vectors is that the distance between them can be calculated much faster. Another advantage is compactness: to store each value one needs only one bit. Unfortunately, many algorithms (including neural networks) provide floating point values and therefore cannot be used directly. In contrast with binary vectors floating point vectors are also used. While the comparison takes more time and storage consumes more space this type of descriptors is still viable because it allows us to employ more algorithms for example neural networks which naturally produce floating point values. Obtaining a neural network binary descriptor is possible but requires additional effort [7], [8] and is a separate problem.

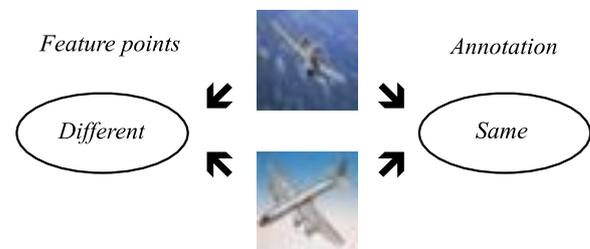

*Fig. 1. Two pictures of a planes. They are the same objects but completely different images*

There is an ambiguity in the task of image description. In paper [9] authors show that published results on different descriptors comparison are inconsistent. But the problem here is even more complex because a set of different tasks is solved using similar methods and algorithms. For example, in [10] authors train an image to vector neural network which is clustering objects representations by their class. It is important that even though the idea of this kind of neural network is the same as for the feature point description, the meaning is completely different. In the first case, two images of a plane should transform into close vectors, and in the second case, the final result must depend on the similarity of the images regardless of the pictured object type as it is demonstrated in Fig. 1. "Image description" problem also exists in even more general form [11].

Existing datasets for the descriptors training are mostly focused on outdoors pictures (for example [12]) and/or too complex to be used for lightweight neural networks training. Moreover, these datasets contain distortions which make them ineffective for document feature point description. To solve this problem we will introduce a new training dataset that is suitable for document feature points descriptor training but can also be used for multiple purposes. In our experiments, we will show that a very lightweight neural network trained on this dataset can show competitive results on both documents and general image patches.

These kinds of neural networks are typically called metric neural networks. To train such a network several loss functions are used. One of them is a triplet loss which was known for a long time already [13]. Despite the fact that this loss function is widely used authors of many papers introduce modifications of the triplet loss for different purposes [14]. Some of them went further for a quadruple loss [15].

To summarise, in this paper we introduce a new training dataset and show how to use it to train a lightweight universal neural network descriptor. The dataset and a method for training data retrieval are provided for public usage.

## 1. Training dataset

### 1.1. Dataset creation

The training dataset consists of five parts which are collected with three different methods: synthetic generation, capturing with a camera, and direct patch generation.

The first part of the dataset contains synthetically generated images with text lines. In document matching templates are usually matched by feature points descriptors. These points are often located on the static texts. So, the final descriptor must evaluate image patches with different letters as different even if these letters are located at the same places. The method described in [16] is used for text printing into the backgrounds which were selected manually.

The second part of the dataset aims for general purposes and consists of various textures collected from the walls, and various random surfaces with a 3D texture. These texture images introduce various shapes and their shades. To ensure the difference between the images we fix the camera and vary light source position like it is demonstrated in Fig. 2.

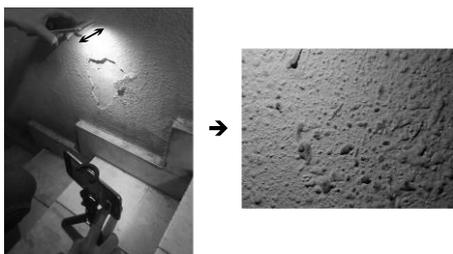

*Fig. 2: Process of the gathering of the second part of the dataset.*

The third part consists of the patches that were generated directly. They are blurry images with intensity peaks in random locations. To achieve this we generated pictures with a white background and several black dots, then apply Gaussian blur and Fast Hough Transform [17] as shown in Fig. 3.

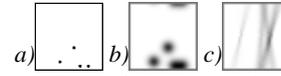

*Figure 3: Patch creation with FHT: initial points, blurred, FHT image.*

In contrast with the rest of the dataset, these images must introduce shapes that are not usually presented in text strings or in the wild. Still, these images are perfectly valid for a patch matching task and therefore a reasonable amount of them will increase the quality of the trained algorithm in general.

The next part is similar to the text strings but instead of letters we used hieroglyphs. This part should cover a big variety of small objects which are not presented in the standard set of symbols.

The final part of the dataset contains images of barcodes instead of letters as they have a lot of small details that the trained algorithm is expected to differentiate.

All these parts of the dataset together contain 85 images of sizes from 1150×388 to 2000×6048. Most of them were duplicated and processed with a graphical editor. By doing this we ensure geometrical matching between the images with the same content and introduce some visual effects which the final descriptor must tolerate.

### 1.2. Patches retrieval

To create a final training set of the patches we process the dataset in a special way (scripts for patches retrieval will be available along with the dataset). Since our neural network is designed to take as input a grayscale picture of the size 32×32 we convert pictures to grayscale and retrieve image patches from all possible positions with small overlapping. This allows us to increase the number of classes and does not mix up classes. To diversify our data and extend the number of classes in the final training data we perform additional steps: add different scales and rotations. Also, we inverse some of the images to further extend the variety of the classes. The exact values of these parameters can be found in the retrieval script.

*Table. 1. Patches per class distribution*

| Patches per class | 1 | 2 | 3 | 4 |
|---|---|---|---|---|
| Classes | 149164 | 138436 | 35624 | 1952 |

Final training data contains 325176 classes and 534860 patches. The distribution of the images per class is shown in Table 1. It is by design that there are many single images per class as we will later employ special data augmentation.

The dataset parts and the amount of classes they yield is summarised in Table 2.

*Table. 2. Classes per data sources*

| Source | Classes number | Samples |
|---|---|---|
| Text lines | 265384 | 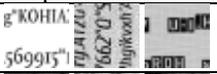 |
| Photos | 38916 | 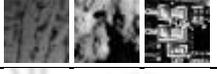 |
| FHT images | 10000 | 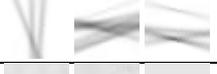 |
| Hieroglyphs | 7140 | 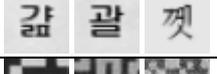 |
| Barcodes | 3736 | 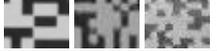 |

Since the designed dataset is created mostly for feature point description on the documents the text lines part is the biggest one in our experimental setup, but it can be balanced using the provided source code.

## 2. Neural network

### 2.1. Architecture

The neural network architecture was created with an extremely small amount of trainable parameters. This architecture has a dimensionality reduction layer (layer 6). This idea is presented in different forms in autoencoders [18], SqueezeNet [19], MobileNets [20] and others. Other than that the neural network is quite simple: the first layer has a 4×4 window size to obtain a noticeable initial receptive field and after that the extracted features are gradually transformed into the final vector with convolutional and fully connected (FC) layers. In the Table 3 we explain the architecture details.

*Table. 3: Neural network architecture. Input shape is 32×32×1*

| # | Layer | Parameters | Activation | Output shape |
|---|---|---|---|---|
| 1 | Conv | 8 filters 4×4, stride 2×2, no padding | symrelu[1] | 15×15×8 |
| 2 | Conv | 8 filters 3×1, stride 1×1, no padding | symrelu[1] | 13×15×8 |
| 3 | Conv | 8 filters 1×2, stride 1×1, no padding | symrelu[1] | 13×13×8 |
| 4 | Conv | 20 filters 3×3, stride 2×2, no padding | symrelu[1] | 6×6×20 |
| 5 | Conv | 16 filters 1×1, stride 1×1, no padding | symrelu[1] | 6×6×16 |
| 6 | Conv | 12 filters 1×1, stride 1×1, no padding | symrelu[1] | 6×6×8 |
| 7 | Conv | 20 filters 2×2, stride 1×1, no padding | symrelu[1] | 5×5×16 |
| 8 | Conv | 48 filters 3×3, stride 2×2, no padding | symrelu[1] | 2×2×48 |
| 9 | FC | 128 outputs | symrelu[1] | 1×1×128 |
| 10 | FC | 16 outputs | - | 1×1×16 |

In this neural network architecture we use ReLU based activation function:

$$symrelu[a] = \max(-a, \min(a, x)), \quad (1)$$

where a>0. This will later allow us to evaluate output value bounds. The resulting neural network has only $3.9*10^4$ trainable parameters which is considered to be very small. For example, HardNet [21] neural networks have much more than $10^6$ parameters. Only $2.4*10^5$ summations and $2.5*10^5$ multiplications are required to evaluate the result which makes this neural network suitable for usage on the device with low computations power such as various smartphones, unmanned vehicles, and others.

### 2.2. Training

#### 2.2.1. Batch generation

To train our neural network we used the patches from a dataset. To generate a training batch we randomly choose 8192 of them. After that, for every patch we also randomly choose one random positive example (i.e. from the same class, if there was only one patch in this class we took the same patch) and one random negative example (i.e. from a different random class). While the current batch is processed by the training framework on GPU we generate the next one on CPU. This part can be improved with various triplet generation techniques like hard mining [22, 23], but this is not a topic of the current paper therefore we use the simplest random selection.

#### 2.2.2. Augmentation

Since our dataset does not have (and was not designed to, see Table 1) many patches for every class, the augmentation part is essential. In our experiments we used an online augmentation system [24]. Image distortions were different for anchor/positive elements and for negative elements of the triplet. For anchor/positive element we carefully select the transformations which should not make the initially similar patches different: monotonic brightness changes, blur, additive noise, random crop and scale, motion blur. For the negative elements, the list of applied transformations was extended with opening and closing morphology operations, grid addition, and highlights. The initial probability of the image augmentation was 0.95. We select a random transformation from the list, apply it to the image with the current probability, then multiply the probability by a factor 0.85 and repeat the procedure until the list is empty. The probability reduction is needed to prevent the data over augmentation. In other words for every image the transformations {T} are shuffled and applied with a probability

$$p(t_i) = 0.95*0.85^i, \quad (2)$$

#### 2.2.3. Loss function and training

After generating and augmenting the batch is passed to a neural network training framework. All the networks were trained for approximately 5000 batches (see Fig. 4 for a convergence plot) before testing. For initial randomization we use Xavier method [25]. All the neural networks were trained with a standard triplet loss function with $\alpha = 1.5$. The convergence plot in Fig. 4 demonstrates interesting behaviour. The blue plot shows the loss and it is decreasing in the training process as it should. The orange plot shows the part of the triplets which were considered to be solved i.e. provided a zero gradient. The green plot shows the part of the triplet where the distance between the anchor and the positive elements was less than $\alpha/2$. We can see that the number of such triplets is decreasing that implies that the representations of the same class grow bigger with time.

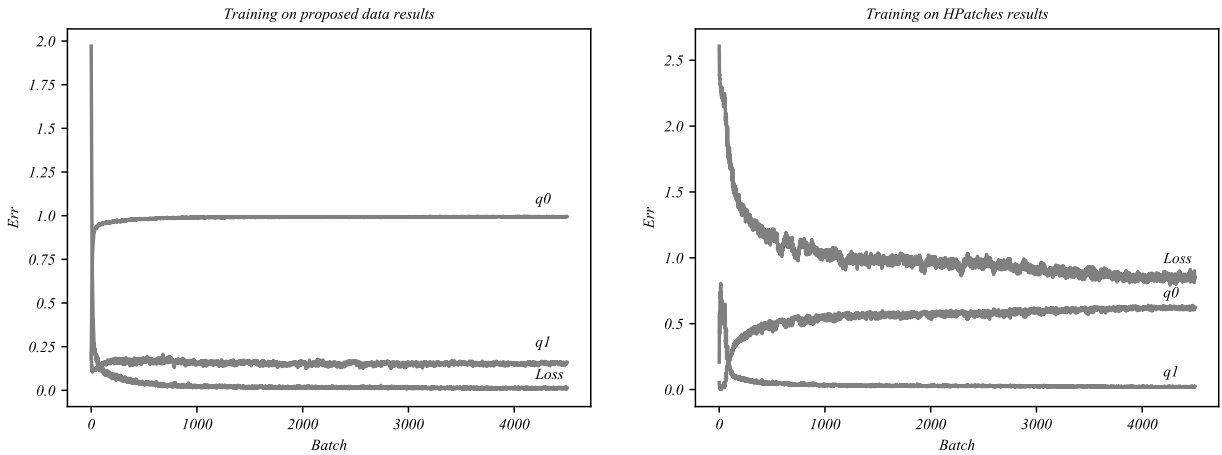

Fig. 4: Neural network convergence plot and training statistics

## 3. Experiments

To prove the effectiveness of our dataset and method we performed four experiments. Firstly we train a neural network on the HPatches training data [9] and on our training data with the original triplet loss function. All neural networks were trained for approximately 5000 batches (each of which consisted of 8192 triplets) with described augmentation.

For testing purposes we used three datasets: HPatches to check the resulting descriptor validity in general and two open datasets containing documents: MIDV-500 [26] and MIDV-2019 [27].

In Fig. 5 we show some images from the used datasets. While HPatches is a dataset of the general image patches mostly containing outdoors images the MIDV-500 and MIDV-2019 datasets contain document images. The second one introduces heavier projective distortions and is considered to be harder. Both datasets have various complex backgrounds and are challenging for the task.

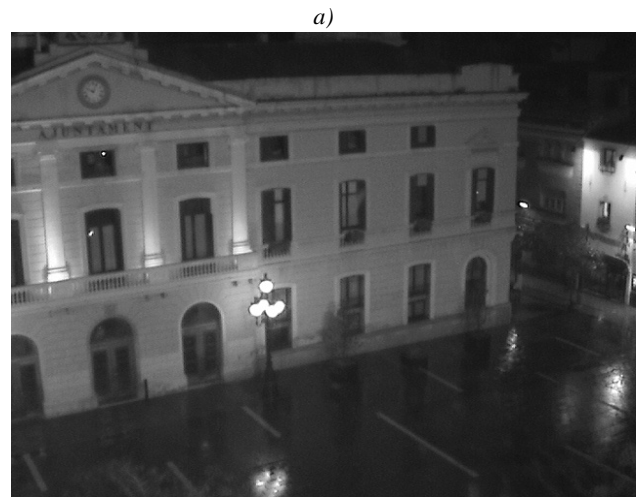

*a)*

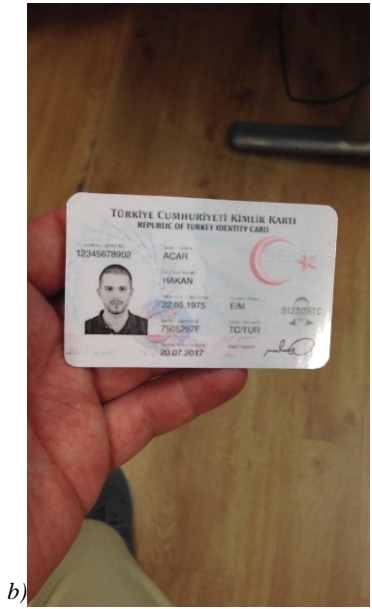

*b)*

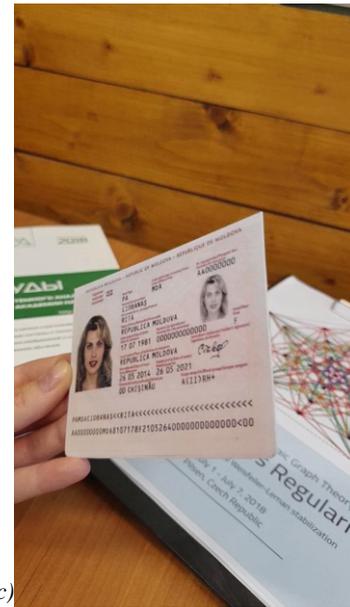

*c)*

*Fig. 5: Examples from the testing datasets: a)HPatches, b)MIDV-500, c)MIDV-2019*

## 4. Results

In Tables 4, 5, and 6 we show the results obtained using HPatches testing framework [9]. It can be seen that in some cases of patch verification (see Tables 4 and 5 our training data were even better. In the retrieval task the situation is even better (see Table 6). The neural network trained on our data shows comparable results.

*Table. 4. Verification task results (balanced). E - easy, H - hard, T - tough*

| Method | E-inter | E-intra | H-inter | H-intra | T-inter | T-intra |
|---|---|---|---|---|---|---|
| ROOTSIFT [28] | 0,904 | 0,874 | 0,799 | 0,762 | 0,715 | 0,680 |
| BRIEF [29] | 0,881 | 0,874 | 0,814 | 0,806 | 0,748 | 0,741 |
| SIFT [30] | 0,931 | 0,910 | 0,823 | 0,796 | 0,731 | 0,705 |
| **Our** | 0,929 | **0,920** | 0,829 | 0,816 | 0,735 | 0,723 |
| Binboost [31] | 0,923 | 0,911 | 0,858 | 0,842 | 0,784 | 0,767 |
| DC-SIAM [32] | 0,933 | 0,910 | 0,875 | 0,843 | 0,813 | 0,776 |
| **Our HP** | 0,936 | 0,904 | 0,893 | 0,846 | 0,847 | 0,791 |
| DC-SIAM2STREAM [32] | 0,951 | 0,934 | 0,920 | 0,896 | 0,874 | 0,842 |
| Deepdesc [33] | 0,959 | 0,936 | 0,931 | 0,896 | 0,888 | 0,842 |
| TFEAT-MARGIN-STAR [34] | 0,963 | 0,947 | 0,937 | 0,913 | 0,894 | 0,861 |
| TFEAT-RATIO-STAR [34] | 0,962 | 0,945 | 0,939 | 0,913 | 0,898 | 0,864 |
| HARDNET [21] | 0,980 | 0,970 | 0,961 | 0,943 | 0,918 | 0,890 |
| HARDNET+ [21] | 0,981 | 0,971 | 0,962 | 0,945 | 0,920 | 0,893 |

*Table. 5. Verification task results (imbalanced). E - easy, H - hard, T - tough*

| Method | E-inter | E-intra | H-inter | H-intra | T-inter | T-intra |
|---|---|---|---|---|---|---|
| ROOTSIFT [28] | 0,778 | 0,695 | 0,582 | 0,484 | 0,455 | 0,367 |
| BRIEF [29] | 0,727 | 0,700 | 0,563 | 0,536 | 0,444 | 0,422 |
| SIFT [30] | 0,849 | 0,783 | 0,657 | 0,570 | 0,512 | 0,429 |
| Binboost [31] | 0,814 | 0,769 | 0,665 | 0,614 | 0,534 | 0,488 |
| **Our** | **0,838** | **0,810** | 0,650 | 0,615 | 0,501 | 0,469 |
| **Our HP** | 0,837 | 0,756 | 0,722 | 0,616 | 0,619 | 0,509 |
| DC-SIAM [32] | 0,845 | 0,785 | 0,724 | 0,644 | 0,609 | 0,524 |
| DC-SIAM2STREAM [32] | 0,884 | 0,841 | 0,806 | 0,745 | 0,705 | 0,632 |
| Deepdesc [33] | 0,904 | 0,851 | 0,830 | 0,751 | 0,733 | 0,640 |
| TFEAT-MARGIN-STAR [34] | 0,916 | 0,874 | 0,846 | 0,781 | 0,748 | 0,668 |
| TFEAT-RATIO-STAR [34] | 0,912 | 0,868 | 0,848 | 0,781 | 0,752 | 0,671 |
| HARDNET [21] | 0,955 | 0,929 | 0,909 | 0,864 | 0,819 | 0,754 |
| HARDNET+ [21] | 0,958 | 0,931 | 0,914 | 0,870 | 0,828 | 0,766 |

*Table. 6. Retrieve task results*

| Method | 100 | 500 | 1000 | 5000 | 10000 | 15000 | 20000 | Mean |
|---|---|---|---|---|---|---|---|---|
| BRIEF [29] | 0,477 | 0,328 | 0,279 | 0,195 | 0,168 | 0,154 | 0,146 | 0,250 |
| **Our HP** | 0,600 | 0,405 | 0,335 | 0,213 | 0,173 | 0,153 | 0,142 | 0,289 |
| **Our** | 0,578 | **0,415** | **0,356** | **0,247** | **0,210** | **0,191** | **0,180** | 0,311 |
| Binboost [31] | 0,575 | 0,416 | 0,363 | 0,269 | 0,235 | 0,218 | 0,208 | 0,326 |
| SIFT [30] | 0,634 | 0,503 | 0,458 | 0,372 | 0,341 | 0,324 | 0,314 | 0,421 |
| ROOTSIFT [28] | 0,625 | 0,501 | 0,460 | 0,384 | 0,355 | 0,340 | 0,331 | 0,428 |
| DC-SIAM2STREAM [32] | 0,709 | 0,562 | 0,509 | 0,399 | 0,360 | 0,339 | 0,326 | 0,458 |
| DC-SIAM [32] | 0,726 | 0,575 | 0,521 | 0,410 | 0,370 | 0,349 | 0,335 | 0,469 |
| TFEAT-RATIO-STAR [34] | 0,737 | 0,602 | 0,549 | 0,437 | 0,397 | 0,375 | 0,361 | 0,494 |
| TFEAT-MARGIN-STAR [34] | 0,745 | 0,614 | 0,564 | 0,455 | 0,415 | 0,394 | 0,380 | 0,510 |
| Deepdesc [33] | 0,774 | 0,644 | 0,587 | 0,469 | 0,427 | 0,403 | 0,388 | 0,527 |
| HARDNET [21] | 0,860 | 0,772 | 0,736 | 0,653 | 0,620 | 0,601 | 0,589 | 0,690 |
| HARDNET+ [21] | 0,861 | 0,773 | 0,738 | 0,655 | 0,623 | 0,605 | 0,593 | 0,693 |

On contrary, on the documents testing datasets we can see, that HPatches training data are not good for the task while our training dataset is suitable. The results from Tables 7 and 8 prove two main points: we need specific training data for documents feature points matching and our data is valid. In these tables "Our HP" shows the neural network trained on HPatches dataset and "Our" shows the result of the one trained on the created dataset.

*Table. 7. MIDV-500 results*

| NN | Total score | Great | Good | Bad | Missing |
|---|---|---|---|---|---|
| Our HP | 0,143 | 1651 | 678 | 2312 | 5336 |
| Our | 0,710 | 9413 | 1621 | 1573 | 1178 |

*Table. 8. MIDV-2019 results*

| NN | Total score | Great | Good | Bad | Missing |
|---|---|---|---|---|---|
| Our HP | 0,059 | 269 | 118 | 547 | 2798 |
| Our | 0,480 | 2436 | 591 | 921 | 1117 |

## 5. Discussion

The suggested neural network has an interesting property: the activation function before the last layer limiting the value. This allows us to evaluate the lower and the upper bounds of the possible neural network output in each dimension. Let consider $W_{i,j}$ and $b_j$ are the weights and biases of the last layer. Then the lower $L$ and the upper $U$ bounds for every dimension can be calculated with equations (3).

$$\begin{cases} L_j = -\sum_i |W_{i,j}| + b_j \\ U_j = \sum_i |W_{i,j}| + b_j \end{cases} \quad (3)$$

Even though we do not use this property in our current work, it can be very useful for the output quantization and reduction of the descriptor size.

Another interesting point is that one can notice that even though for the triplet training multiple examples per class are needed to construct anchor positive occurrences in our data there are many classes with a single image. It may seem that this is a disadvantage of the dataset but in fact on contrary. With this data distribution we can carefully choose augmentation for anchor and positive image and control which transformations should be tolerated and which should be not.

Finally, since most of the data is randomly generated we cannot be 100% sure that there are no images in different classes that are actually very similar. But our analysis shows that the probability of this is very low. Furthermore, with over $3*10^5$ classes the chance, that two exact images will be selected incorrectly is negligible.

### *Conclusion*

In this paper, we showed that feature points description is different for documents and for outdoor images. The comparison of the trained neural networks on the general and special (ours) datasets clearly shows the gap in the quality. The main purpose of our dataset is to provide the necessary information for the description of the image patches containing letters. Additional images make the training data applicable not only for document image patches matching but for other purposes as well. We also demonstrated that a very lightweight neural network can still be used for the task which makes this kind of algorithm applicable when using on devices with limited computational resources.

For future work we plan to further enhance the dataset in two main different ways: evaluate what type of data is still missing and add new images and improve the patches retrieval mechanism to use the already presented data even more efficiently. Also, we plan to quantize the neural network and its output down from 32 bits per value to 8 bits per value which should be possible without (or with minimal) quality loss according to the neural network properties. We also plan to study the possibility of the output dimension reduction for further descriptor size shrink.

### *References*

[1] Kougia V, Pavlopoulos J, Androutsopoulos I. Medical Image Tagging by Deep Learning and Retrieval. In: Arampatzis A. et al. (eds) Experimental IR Meets Multilinguality,

Multimodality, and Interaction. CLEF 2020. Lecture Notes in Computer Science, vol. 1260, pp. 154–166. Springer, Cham (2020). 10.1007/978-3-030-58219-7_14

[2] Shin Y, Seo K, Ahn J, Im DH. Deep-Learning-Based Image Tagging for Semantic Image Annotation. In: Park J., Park DS., Jeong YS., Pan Y. (eds) Advances in Computer Science and Ubiquitous Computing. CUTE 2018, CSA 2018. Lecture Notes in Electrical Engineering, vol 536, pp. 54–59. Springer, Singapore (2019). 10.1007/978-981-13-9341-9_10

[3] William I, Ignatius Moses Setiadi DR, Rachmawanto EH, Santoso HA, Sari CA. Face Recognition using FaceNet (Survey, Performance Test, and Comparison). In: Fourth International Conference on Informatics and Computing 2019, ICIC, vol. 1, pp. 1–6. Semarang, Indonesia (2019). 10.1109/ICIC47613.2019.8985786

[4] Skoryukina N, Arlazarov V, Nikolaev D. Fast method of ID documents location and type identification for mobile and server application. In: International Conference on Document Analysis and Recognition, 2019, ICDAR, vol. 1, pp. 850–857. IEEE Xplore, Sydney (2019). 10.1109/ICDAR.2019.00141

[5] Kumar M, Gupta S, Mohan N. A computational approach for printed document forensics using SURF and ORB features. Soft Comput 24(1), 13197–13208 (2020). 10.1007/s00500-020-04733-x

[6] Ilyuhin SA, Sheshkus AV, Arlazarov VL. Recognition of images of Korean characters using embedded networks. In: Wolfgang, O., Nikolaev D., Zhou J. (eds.) Twelfth International Conference on Machine Vision 2019, ICMV, vol. 11433, pp. 1–7. SPIE, Amsterdam (2020). 10.10007/1234567890

[7] Duan Y, Lu J, Wang Z, Feng J, Zhou J. Learning Deep Binary Descriptor with Multi-quantization. In: IEEE Conference on Computer Vision and Pattern Recognition 2017, CVPR, vol. 1, pp. 1183–1192. IEEE Xplore, Honolulu USA (2017). 10.1109/CVPR.2017.516

[8] Zhang J, Ye S, Huang T, Rui Y. CDbin: Compact Discriminative Binary Descriptor Learned With Efficient Neural Network. IEEE Transactions on Circuits and Systems for Video Technology 30(3), 862–874 (2020). 10.1109/TCSVT.2019.2896095

[9] Balntas V, Lenc K, Vedaldi A, Mikolajczyk K. HPatches: A benchmark and evaluation of handcrafted and learned local descriptors. In: IEEE conference on computer vision and pattern recognition 2017, CVPR, pp. 5173–5182. Springer, Heidelberg (2017).

[10] Hoffer E, Ailon N. Deep Metric Learning Using Triplet Network. In: Feragen A., Pelillo M., Loog M. (eds) Similarity-Based Pattern Recognition 2015, SIMBAD, vol. 9370, pp. 84–92. Springer, Cham (2015). 10.1007/978-3-319-24261-3_7

[11] Mishra A, Liwicki M. Using deep object features for image descriptions. arXiv preprint arXiv:1902.09969. 1–5 (2019)

[12] Paulin M, Douze M, Harchaoui Z, Mairal J, Perronin F, Schmid C. Local convolutional features with unsupervised training for image retrieval. In: Editor,F., Editor, S. (eds) Proceedings of the IEEE international conference on computer vision 2015, ICCV, vol. 1, pp. 91–99. IEEE Xplore, Santiago (2016). 10.1109/ICCV.2015.19

[13] Schultz M, Joachims T. Learning a distance metric from relative comparisons. Advances in neural information processing systems 16(1), 41–48 (2004).

[14] Cacheux YL, Borgne HL, Crucianu M. Modeling inter and intra-class relations in the triplet loss for zero-shot learning. In: Proceedings of the IEEE/CVF International Conference on Computer Vision 2019, ICCV, vol. 1, pp. 10333–10342. IEEE Xplore (2019).

[15] Chen W, Chen X, Zhang J, Huang K.: Beyond triplet loss: a deep quadruplet network for person re-identification. In: Proceedings of the IEEE conference on computer vision and pattern recognition 2017, CVPR, vol. 1, pp. 403–412. IEEE Xplore (2017).

[16] Chernyshova YS, Gayer AV, Sheshkus AV. Generation method of synthetic training data for mobile OCR system. In: Tenth international conference on machine vision 2017, ICMV, vol. 10696, pp. 106962G. SPIE, Vienna (2018). 10.1117/12.2310119

[17] Nikolaev DP, Karpenko SM, Nikolaev IP, Nikolayev PP. Hough transform: underestimated tool in the computer vision field. In: Louca, L. S., Chrysanthou, Y., Oplatkova, Z., Al-Begain, K. (eds.) Proceedings of the 22th European Conference on Modelling and Simulation 2008, ECMS, vol. 238, pp. 238–243. European Council for Modeling and Simulation, Nicosia (2016). 10.7148/2008-0238

[18] Hinton GE, Salakhutdinov RR. Reducing the Dimensionality of Data with Neural Networks. Science 313(5786), 504–507 (2006). 10.1126/science.1127647

[19] Iandola FN, Han S, Moskewicz MW, Ashraf K, Dally WJ, Keutzer K. SqueezeNet: AlexNet-level accuracy with 50x fewer parameters and< 0.5 MB model size. arXiv preprint arXiv:1602.07360, 1–13 (2016).

[20] Howard AG, Zhu M, Chen B, Kalenichenko D, Wang W, Weyand T, Andreetto M, Adam H. Mobilenets: Efficient convolutional neural networks for mobile vision applications. arXiv preprint arXiv:1704.04861, 99–110 (2017).

[21] Mishchuk A, Mishkin D, Radenovic F, Matas J. Working hard to know your neighbor's margins: Local descriptor learning loss. In: Advances in Neural Information Processing Systems 30 (NIPS 2017), pp. 4826–4837. Curran Associates, Montreal (2017).

[22] Zhao Y, Jin Z, Qi GJ, Lu H, Hua XS. An adversarial approach to hard triplet generation. In: Ferrari V., Hebert M., Sminchisescu C., Weiss Y. (eds) Computer Vision – Proceedings of the European conference on computer vision 2018, ECCV, vol. 11213, pp. 501–517. Springer, Cham (2018). 10.1007/978-3-030-01240-3_31

[23] Sikaroudi M, Ghojogh B, Safarpoor A, Karray F, Crowley M, Tizhoosh HR. Offline Versus Online Triplet Mining Based on Extreme Distances of Histopathology Patches. In: Bebis G. et al. (eds. ) Advances in Visual Computing 2020, ISVC, vol. 12509, pp. 333–345. Springer, Cham (2020). 10.1007/978-3-030-64556-4_26

[24] Gayer AV, Chernyshova YS, Sheshkus AV. Effective real-time augmentation of training dataset for the neural networks learning. In: Eleventh International Conference on Machine Vision 2018, ICMV, vol. 11041, pp. 104111. SPIE, Munich (2019). 10.1117/12.2522969

[25] Glorot X, Bengio Y. Understanding the difficulty of training deep feedforward neural networks. In: Yee Whye Teh, Titterington, M. (eds) Proceedings of the thirteenth international conference on artificial intelligence and statistics 2010, AISTAST, vol. 9, pp. 249–256. JMLR Workshop and Conference Proceedings (2010).

[26] Arlazarov VV, Bulatov K, Chernov T, Arlazarov VL. MIDV-500: A Dataset for Identity Document Analysis and Recognition on Mobile Devices in Video Stream. Computer Optics 43(5), 818–824 (2019). 10.18287/2412-6179-2019-43-5-818-824

[27] Bulatov K, Matalov D, Arlazarov VV. MIDV-2019: challenges of the modern mobile-based document OCR. In:

Twelfth International Conference on Machine Vision 2019, ICMV, vol. 11433, pp. 114332N. SPIE, Munich (2019). 10.1117/12.2558438
[28] Arandjelovic R, Zisserman A. Three things everyone should know to improve object retrieval. In: Proceedings of the 2012 IEEE Conference on Computer Vision and Pattern Recognition, pp. 2911-–2918. IEEE, Rhode Island, USA (2012). 10.1109/CVPR.2012.6248018
[29] Calonder M, Lepetit V, Strecha C, Fua P. BRIEF: Binary robust independent elementary features. In: Proceedings of the 11th European Conference on Computer Vision, pp. 778-–792. Springer, Berlin, Heidelberg (2010). 10.1007/978-3-642-15561-1_56
[30] Lowe DG. Object recognition from local scale-invariant features. In: Proceedings of the Seventh IEEE International Conference on Computer Vision, vol. 2, pp. 1150-–1157. IEEE, Kerkyra, Greece (1999). 10.1109/ICCV.1999.790410
[31] Trzcinski T, Christoudias M, Lepetit V. Learning image descriptors with boosting. In: IEEE Transactions on Pattern Analysis and Machine Intelligence, 37(3), pp. 597-–610, 2015. 10.1109/TPAMI.2014.2343961
[32] Zagoruyko S, Komodakis N. Learning to compare image patches via convolutional neural networks. In: Proceedings of the 2015 IEEE Conference on Computer Vision and Pattern Recognition (CVPR), pp. 4353-4361. IEEE, Boston, MA, USA (2015). 10.1109/CVPR.2015.7299064
[33] Simo-Serra E, Trulls E, Ferraz L, Kokkinos I, Fua P, Moreno-Noguer F. Discriminative learning of deep convolutional feature point descriptors. In: Proceedings of the 2015 IEEE International Conference on Computer Vision (ICCV), pp. 118–126. IEEE, Santiago, Chile (2015). 10.1109/ICCV.2015.22
[34] Balntas V, Riba E, Ponsa D, Mikolajczyk K. Learning local feature descriptors with triplets and shallow convolutional neural networks. In: Proceedings of the British Machine Vision Conference, pp. 119.1–119.11. BMVA Press (2016). 10.5244/C.30.119
[35] Pultar M. Improving the HardNet Descriptor. Czech Technical University in Prague. Faculty of Electrical Engineering Department of Cybernetics. 1(1), 1–48 (2020).

*Author's information*

**Alexander Vladimirovich Sheshkus** (b. 1986) received the B.S. and M.S. degrees in Applied Physics and Mathematics from Moscow Institute of Physics and Technology (State University), Moscow, Russia, in 2009 and 2011, respectively. He is currently the head of Machine Learning department in Smart Engines, and a researcher in FRC "Computer Science and Control" of RAS. His research interests include deep neural networks, computer vision and projective invariant image segmentation. E-mail: *astdcall@gmail.com*.

**Anastasiya Nikolaevna Chirvonaya** (b. 1998) received the B.S. degree in Applied Mathematics from National University of Science and Technology "MISIS", Moscow, Russia, in 2019. She is currently pursuing the M.S. degree in Applied Science at the same university. She works as a programmer at Smart Engines. Her research interests are computer vision and machine learning. E-mail: *nastyachirvonaya@smartengines.biz*.

**Vladimir Lvovich Arlazarov** (b. 1939) Dr. Sc., corresponding member of the Russian Academy of Sciences, graduated from Lomonosov Moscow State University in 1961. Currently he works as head of sector 9 at the Institute for Systems Analysis FRC "Computer Science and Control" RAS. Research interests are game theory and pattern recognition. E-mail: *vladimir.arlazarov@gmail.com*.